%% file: root.tex
\documentclass[letterpaper, 10 pt, conference]{ieeeconf}  

\IEEEoverridecommandlockouts                              

\usepackage{color}
\usepackage{pgfplots}
\usepackage{url}
\usepackage{multirow}
\usepackage{graphicx}

\makeatletter
\let\NAT@parse\undefined
\makeatother

\usepackage{hyperref}

\title{\LARGE \bf
Jacquard: A Large Scale Dataset for Robotic Grasp Detection
}

\author{Amaury Depierre$^{1,2}$, Emmanuel Dellandr\'ea$^{2}$ and Liming Chen$^{2}$
\thanks{$^{1}$Sil\'eane, Saint-Etienne, France
        {\tt\small a.depierre@sileane.com}}%
\thanks{$^{2}$University of Lyon, Ecole Centrale de Lyon, LIRIS, CNRS UMR 5205, France
        {\tt\small \{emmanuel.dellandrea, liming.chen\}@ec-lyon.fr}}%
\thanks{Preprint version, accepted at IEEE/RSJ IROS 2018}
}

\begin{document}

\maketitle
\thispagestyle{empty}
\pagestyle{empty}

\begin{abstract}

Grasping skill is a major ability that a wide number of real-life applications require for robotisation. State-of-the-art robotic grasping methods perform prediction of object grasp locations based on deep neural networks. However, such networks require huge amount of labeled data for training making this approach often impracticable in robotics. In this paper, we propose a method to generate a large scale synthetic dataset with ground truth, which we refer to as the Jacquard grasping dataset. Jacquard is built on a subset of ShapeNet, a large CAD models dataset, and contains both RGB-D images and annotations of successful grasping positions based on grasp attempts performed in a simulated environment. We carried out experiments using an off-the-shelf CNN, with three different evaluation metrics, including real grasping robot trials. The results show that Jacquard enables much better generalization skills than a human labeled dataset thanks to its diversity of objects and grasping positions. For the purpose of reproducible research in robotics, we are releasing along with the Jacquard dataset a web interface for researchers to evaluate the successfulness of their grasping position detections using our dataset.


\end{abstract}

\section{INTRODUCTION}

Despite being a very simple and intuitive action for a human, grasp planning is quite a hard task for a robotic system. Detecting potential grasp for a parallel plate gripper from images involves segmenting the image into objects, understanding their shapes and mass distributions and eventually sending coordinates to the robot's actuator. As the whole trajectory of the arm and its end position depend on these coordinates, precision is critical and an error of one pixel in the prediction can make the difference between success and failure of the grasping. Because of these difficulties and despite the progress made recently, performance for this task is still far from what we could expect for real-case applications.

State-of-the-art methods to predict a grasping position for a parallel plate gripper from visual data rely on deep neural networks trained either to directly predict a grasp \cite{redmon2015real} or to evaluate the quality of previously generated candidates and select the best one \cite{mahler2017dex}. These methods rely on supervised training based on labeled data, which may be obtained through one of the following techniques: human labeling, physical trials with robots \cite{pinto2016supersizing} \cite{levine2016learning}, analytic computation where a model is used to predict the effect of external forces applied on the model \cite{rodriguez2012caging} and physics simulation for which the grasp is performed in a computer simulation of the real world \cite{miller2004graspit}. The first two methods, despite being the most accurate, are very time-consuming and therefore cannot be easily used to generate very large datasets. The last two, on the other hand, can be used quite easily to generate millions of labeled data, but generally require to match the CAD model to the position of the object in the image to be efficient.

\begin{figure}
	\centering
	\includegraphics[width=0.45\textwidth]{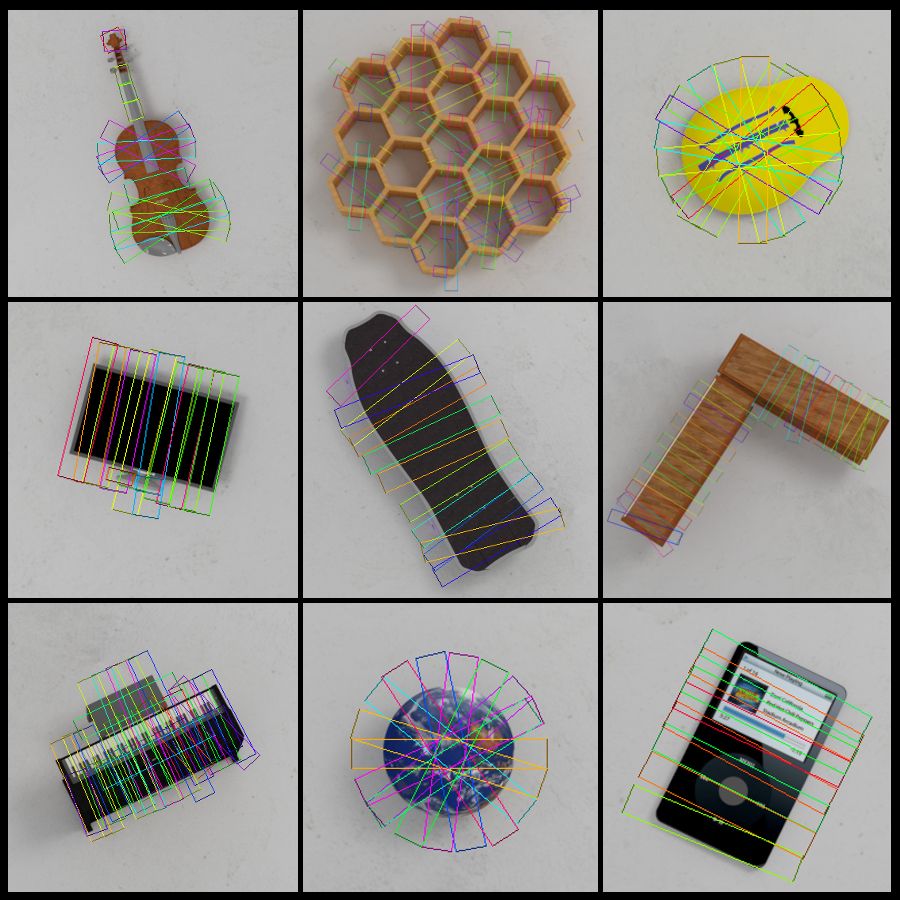}
	\caption{Jacquard dataset contains a large diversity of objects, each with multiple labeled grasps on realistic images. Grasps are drawn as 2D rectangles on the image, darker sides indicate the position of the jaws.}
	\label{dataset_example}
\end{figure}

In this paper, we present an approach to automatize the generation of labeled images for robotic grasping by simulating an environment as close as possible of a physical setup. With this environment, we created the Jacquard dataset containing more than one million unique grasp locations on a large diversity of objects. \hyperref[dataset_example]{Fig. 1} shows some examples of annotated images from Jacquard dataset. We also introduce a novel criterion, namely simulated grasp trial (SGT), to judge the goodness of a grasp location prediction based on physic grasp simulation. This criterion comes in contrast to the distance-based metrics traditionally used for the evaluation of grasp prediction and sticks with the fact that a single object can depict a large number of grasping positions, including those which are not necessarily previously annotated. Using three different evaluation metrics, including SGT and assessment through a real grasping robot trials, we show that this novel dataset, despite being  synthetic, can be used to train a deep neural network (DNN), for grasp detection from a simple image of the scene and achieve much better prediction of grasp locations, in particular for unseen objects, than the same DNN when it is trained using a grasp dataset with human labeled data.

This paper is organized as follows. Section II overviews the related work. Section III states the modelisation we used to describe a grasp. Section IV presents the method used to generate Jacquard dataset. Section V discusses the experimental results using the Jacquard dataset in comparison with the Cornell dataset. Section VI concludes the paper.

\section{RELATED WORK}

Early research in grasp prediction assumed the robot to have a perfect knowledge of its environment and aimed to plan grasps based on a 3D model of the object \cite{miller2003automatic} \cite{bohg2010learning}. Using this technique, Goldfeder et al. \cite{goldfeder2009columbia} created the Columbia Grasp Database, containing more than 230k grasps. With this type of approach, the notion of image is not present, only the CAD models of the gripper and  objects are used. At test time, a query object is matched with an object within the database and a grasp is generated using the similarity of the CAD models. With this approach, both the model and the position of the object have to be known at test time, which is generally not the case for real-world applications.

Recent development of deep learning \cite{lecun2015deep} and more particularly of the Convolutional Neural Networks (CNN) have inspired many researchers to work directly on images instead of 3D CAD models. The simultaneous apparition of cheaper sensors as the Kinect, also helped by providing additional depth information to the RGB image. This led to the development of datasets based on physical trials. In \cite{oberlin15scale} a method to share the knowledge of different robots was developed in order to collect a large collection of data, in \cite{pinto2016supersizing} a Baxter robot has been used to collect 50k data, while in \cite{levine2016learning} the authors collected over 800k datapoints (image, grasp position and outcome) using 14 robotic arms running during two months. In the last two cases, a CNN was successfully trained to detect grasp positions from the collected data. However, these approaches are either material or time consuming or and can not be fully automatized: human intervention is still needed to position the objects in front of the robot. Moreover, these methods only generate one single grasp annotation whereas there are generally several positions which could be good for robotic grasping.

To overcome the issue of time-consuming data generation, Mahler et al. \cite{mahler2017dex} created Dexnet-2.0, a synthetic dataset with 6.7 millions depth images annotated with the success of the grasp performed at the center of the image. They trained a Grasp Quality CNN with these data and achieved a 93\% success rate when predicting the outcome of a grasp. The GQ-CNN has good performance, but it can not be trained end-to-end to predict grasp positions: it only takes grasp candidates generated by another method as an input and rank them.

In \cite{johns2016deep}, Johns et al. used a similar approach: they simulated grasp attempts on 1000 objects and trained a neural network to predict a score over a predefined grid of possible positions for the gripper. The network's input was a depth image, but they did not release their data publicly. In comparison, our Jacquard dataset contains more than 11k objects with both RGB and realistic depth information created through stereo-vision.

The dataset most similar to our work is the Cornell Grasping Dataset \footnote{\url{http://pr.cs.cornell.edu/grasping/rect_data/data.php}}. It is composed of 885 RGB-D images of 240 different objects with 8019 hand-labeled grasp rectangles. As shown in \cite{redmon2015real}, this dataset enables the training of a neural network to detect grasps in images. However a dataset with 885 images is quite small compared to the ones traditionally used in deep learning and may lead to bad performance when generalizing on different images or object configurations. Human labeling can also be biased to grasps that are easily performed with a human hand but not necessarily with a parallel plate gripper. Comparatively, the proposed Jacquard dataset is more than 50 times bigger with various objects and grasps sizes and shapes. A summary of the properties of public grasp datasets can be found in \hyperref[table_1]{Table I}.

\begin{table*}[]
\centering
\caption{Summary of the properties of publicly available grasp datasets}
\label{table_1}
\begin{tabular}{|l|c|c|c|c|c|c|c|c|}
\hline
\multicolumn{1}{|c|}{\textbf{Dataset}} & \textbf{\begin{tabular}[c]{@{}c@{}}Number of\\ objects\end{tabular}} & \textbf{Modality} & \textbf{\begin{tabular}[c]{@{}c@{}}Number of\\ images\end{tabular}} & \textbf{\begin{tabular}[c]{@{}c@{}}Multiple gripper\\sizes \end{tabular}} & \textbf{\begin{tabular}[c]{@{}c@{}}Multiple grasps\\ per image \end{tabular}} & \textbf{\begin{tabular}[c]{@{}c@{}}Grasp\\location\end{tabular}} & \textbf{\begin{tabular}[c]{@{}c@{}}Number of\\grasps\end{tabular}} & \textbf{\begin{tabular}[c]{@{}c@{}}Automatized\\ generation\end{tabular}} \\ \hline
Levine et al. \cite{levine2016learning} & - & RGB-D & 800k & No & No & Yes & 800k & No \\ \hline
Mahler et al. \cite{mahler2017dex} & 1500 & Depth & 6.7M & No & No & No & 6.7M & Yes \\ \hline
Cornell & 240 & RGB-D & 1035 & Yes & Yes & Yes & 8019 & No \\ \hline
Jacquard (ours) & 11k & RGB-D & 54k & Yes & Yes & Yes & 1.1M & Yes \\ \hline
\end{tabular}
\end{table*}

\section{MODELLING ROBOTIC GRASP}
%
%
\label{sec:notations}

\begin{figure}
	\centering
	\def\svgwidth{0.45\textwidth}
	\LARGE
	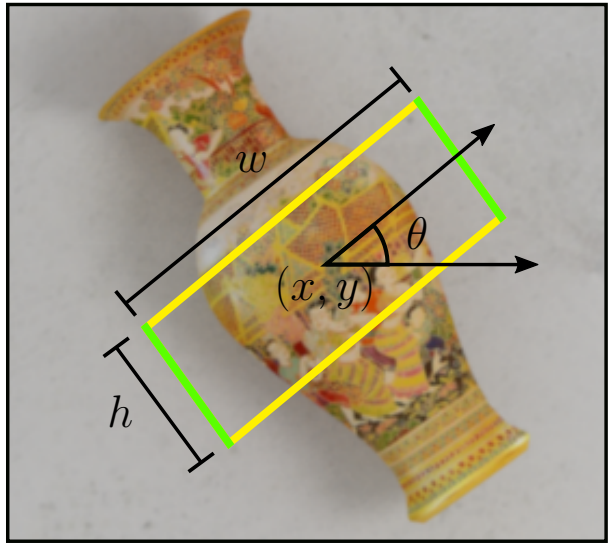
	\caption{Parametrization of a grasp for a parallel-plate gripper. A grasp is described as a five dimensional vector: two values for the position of the center, two for its size and one for its orientation with respect to the horizontal axis. Green sides represent the inner sides of the parallel jaws, yellow sides show the opening of the gripper.}
	\label{grasp_notations}
\end{figure}

In this work, we are  interested in finding a good grasp from a RGB-D image of a single object laying on a plane. A grasp is considered good when the object is successfully lifted and moved away from the table by a robot with a parallel-plate gripper. As shown in \hyperref[grasp_notations]{Fig. 2}, a grasp can be described as:
\begin{equation}
g=\{x,y,h,w,\theta\}
\end{equation}

where $(x,y)$ is the center of a rectangle, $(h,w)$ its size and $\theta$ its orientation relative to the horizontal axis of the image. This representation differs from the one of seven dimensions described in \cite{jiang2011efficient} but Lenz et al. show in \cite{lenz2015deep} that it works well in practice. The main advantage of this representation is that the grasp can be simply expressed  in the image coordinates' system, without any information about the physical scene: $z$ position of the parallel plates and approach vector are determined from the depth image.
When the grasp is performed by a real robot, $h$ and $w$ are respectively fixed and bounded by the shape of the gripper.

\section{JACQUARD DATASET}

\begin{figure*}
	\centering
	\def\svgwidth{\textwidth}
    \small
	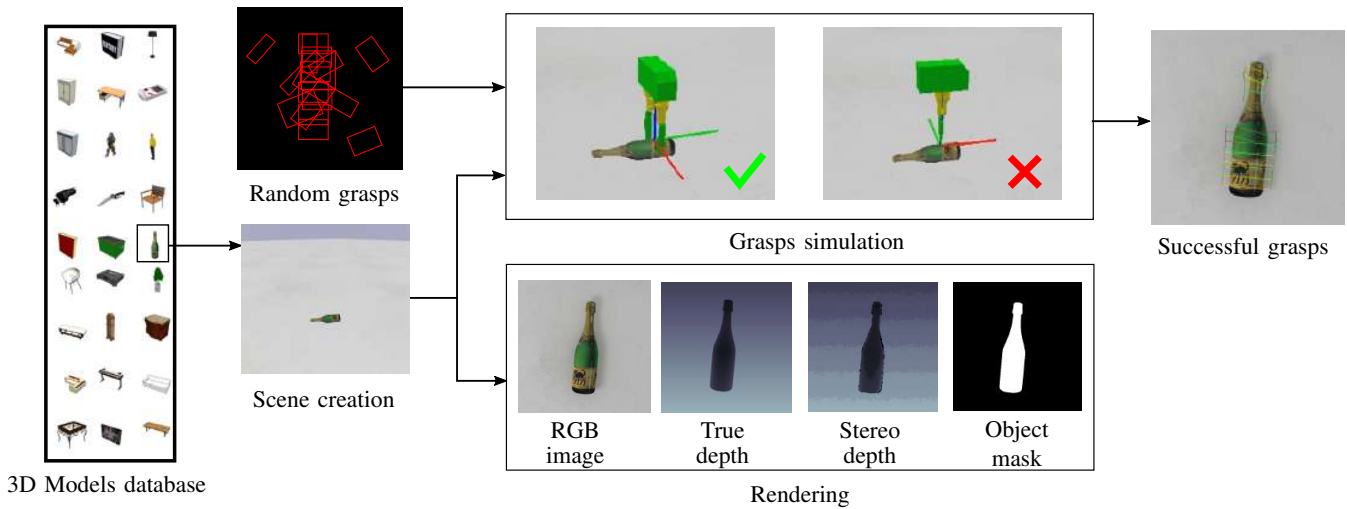
	\caption{The pipeline we used to generate annotated images from 3D models. Random grasps are generated from a probability map obtained with a simple heuristic algorithm before being tested in the simulation environment. In the rendering part, a synthetic camera renders the different images.}
	\label{generation_pipeline}
\end{figure*}

To solve the problem of data starvation, we designed a new method to get images and ground truth labels from CAD models through simulation. Then we applied this process to a subset of ShapeNet \cite{shapenet2015}, namely ShapeNetSem \cite{savva2015semgeo}, resulting in a new dataset with more than 50k images of 11k objects and 1 million unique successful grasp positions annotated. These data are made available to the research community \footnote{\url{http://jacquard.liris.cnrs.fr}}. The main pipeline we used for data generation is illustrated on \hyperref[generation_pipeline]{Fig. 3}. Physics simulation were performed using pyBullet library \cite{coumans2017} and Blender \cite{blender2016} was used to render the images through its Cycles Renderer.

\subsection{Scene creation}

Scenes are all created in the same way. A plane with a white texture is created, the texture being randomly rotated and translated to avoid constant background. Then we select an object from a pool of CAD models. As the objects in ShapeNet have a wide range of scales, we rescale the model so the longest side of its bounding box has a length between 8 and 90 cm. We also give the object a mass depending on its size (80 g for a 8 cm object and 900 g for a 90 cm one) before dropping it from a random position and orientation above the plane. Once the object is in a stable position, the scene configuration is saved.

This scene description is sent to two independent modules: one to render the images and one with a physics simulator to generate the grasp annotations. For the Jacquard dataset, we created up to five scenes for each object. This number was chosen in order to have different views of the objects, but can be increased without any change in the process if necessary.

\subsection{Image rendering}

RGB and true depth images are rendered with Blender. To stick as close as possible to real scene images, instead of adding Gaussian noise to the perfect depth image as in \cite{johns2016deep}, we rendered two more RGB synthetic images with a projected pattern and applied an off-the-shelf stereo-vision algorithm \cite{hirschmuller2008stereo} on them, giving a noisy depth. This approach has been shown to produce depth images very close to real ones in \cite{bregier2017symmetry}. A binary mask separating the object and the background is also created.

\subsection{Annotation generation}

To generate grasp annotations, we used the real-time physics library pyBullet. As for the rendering module, the object model is loaded into the pyBullet environment, however, to speed up calculations, collisions are not computed directly on the mesh but on a volumetric hierarchical approximate convex decomposition \cite{vhacd16} of it. Different grippers with parallel-jaws are simulated. They all have a max opening of 10 cm and a jaw size in $\{1, 2, 3, 4, 6\}$ cm. The different jaw sizes for the gripper combined with the varied scales of  objects ensure that our simulated gripper can perform grasps in a wide range of different configurations.

Grasp annotations are generated in three steps. First, we generate thousands of random grasp candidates covering the whole area under the camera. Then, all these grasp candidates are tested through rigid body simulation using a gripper with a jaw size of 2 cm. And finally all the successful positions of the previous step are tested again with all the gripper sizes. The result is a set of successful grasp locations, each having between 1 and 5 jaw sizes.

To perform simulated grasps, the approach vector is set to the normal at the center of the grasp and the orientation and opening of the gripper are defined by the rectangle coordinates as described in section \ref{sec:notations}. A grasp is considered successful if the object is correctly lifted, moved away and dropped at a given location by the simulated robot. Once all the random candidates have been tested, a last pass is performed on good grasps to remove the ones which are too close from each other. This last step is necessary to ensure that all the grasps are annotated only once.

As the number of possible grasps for one image is very large, we used a non-uniform probability distribution: candidates are generated more frequently in the most promising areas. Theoretically, candidates could be generated with a uniform distribution, but in this case many grasps would fall in an empty area without the object. For the Jacquard dataset, we used a simple heuristic looking for aligned edges in the image and generating the probability distribution from the density of such edges. However, our experiments showed us that any reasonable heuristic lead to a similar final grasps distribution in the image, at the cost of more random trials. With this method, we can reduce the number of grasp attempts necessary to annotate a scene by orders of magnitudes, while keeping a diversity in grasp locations. Such a diversity is very important for deep learning oriented methods.

\subsection{Assessment criterion of successful grasp predictions}
\label{sec:criterion}

With the Cornell Grasp Dataset, the criterion used to determine whether a grasp prediction is correct or not is a rectangle-based metrics. With this criterion, a grasp is considered to be correct if both:
\begin{itemize}
  \item The angle between the prediction and the ground-truth grasp is smaller than a threshold (a typical value is 30$^{\circ}$)
  \item The intersection over union ratio between the prediction and the ground-truth grasp is over a threshold (typically 25\%)
\end{itemize}

\begin{figure}
	\centering
	\def\svgwidth{0.45\textwidth}
	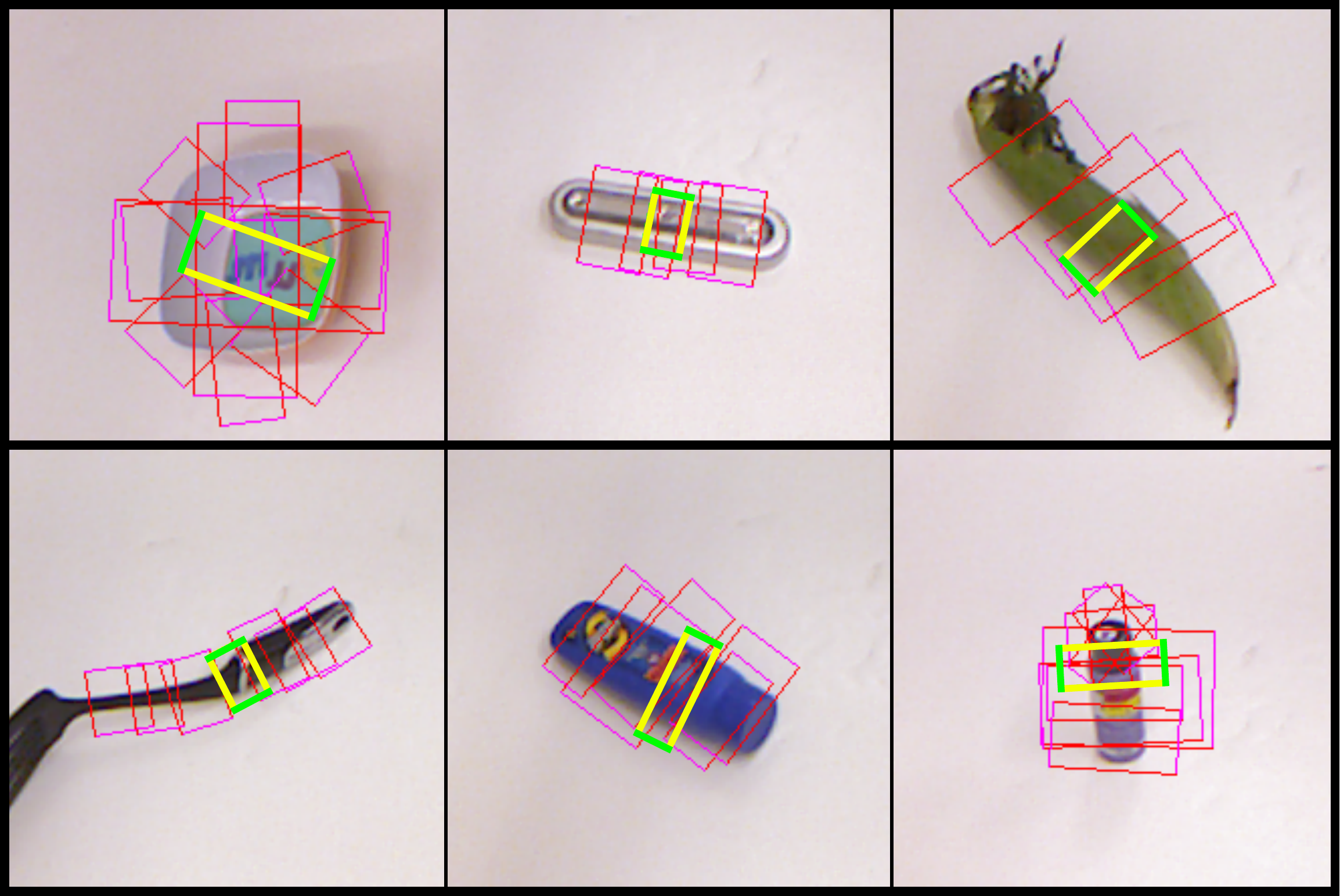
	\caption{Example of misclassifications with the rectangle metrics. Prediction is in yellow and green, ground truth is in red and purple. Top row shows false positives, bottom row shows false negatives.}
	\label{misclassification}
\end{figure}

This criterion can however produce a lot of ``visually'' false-positives, \textit{i.e.}, grasps that, from our human expertise, look bad, but that  the rectangle metrics predict as good, as well as false-negatives, \textit{i.e.}, grasps that, from our human expertise, look  good, but that  the rectangle metrics predict bad. \hyperref[misclassification]{Fig. 4} shows some examples of such misclassifications.

With the Jacquard dataset, we propose a new criterion based on simulation, subsequently called simulated grasp trial-based criterion (SGT). Specifically, when a new grasp should be evaluated as successful or not, the corresponding scene is rebuilt in the simulation environment and the grasp is performed by the simulated robot, in the exact same conditions as during the generation of the annotations. If the outcome of the simulated grasp is a success, \textit{i.e.}, the object is successfully lifted and moved away by the simulated robot using the predicted grasp location, the prediction is then considered as a good grasp. This novel SGT criterion is  much closer than the rectangle metrics to real-world situations where a single object can have many successful grasp locations, including successful grasp locations which are not previously annotated.  
For the purpose of reproducible research, we are releasing along with the dataset a web interface allowing researchers to send grasp requests on our simulator and receive the corresponding grasp outcome.

\section{EXPERIMENTS AND RESULTS}
\label{sect:experiments and results}

In order to evaluate the effectiveness of the proposed simulated Jacquard grasp dataset, we carried out two series of experiments: 1) cross-dataset grasp prediction with the Cornell and Jacquard datasets (section \ref{subsect:cross-dataset evaluation}); 2) evaluation of grasp predictions using a real grasping robot (section \ref{subsect:Evaluation of grasp predictions using a real grasping robot}). We start by explaining the training setup. 

\subsection{Training setup}
\label{subsect:Experimental setup}

In all our experiments, we used an off-the-shelf CNN, \textit{i.e.}, AlexNet\cite{krizhevsky2012imagenet}. The network's convolution weights have been pre-trained  on ImageNet \cite{imagenet_cvpr09} while the fully connected layers are trained from scratch. To use AlexNet with RGB-D, we simply normalize the depth image to get values close to color channels and duplicate the blue filters in the first pre-trained convolution layers. The network is trained through Stochastic Gradient Descent algorithm for 100k iterations with a learning rate of 0.0005, a momentum of 0.9 and a weight decay of 0.001. The learning rate is set to 0.00005 after the first 75k iterations. To compute the error of the network, the Euclidean distance between the prediction and the closest annotation is used:

\begin{equation}
\mathcal{L}=\min_{g \in \mathcal{G}} {\Vert g -  \hat{g} \Vert}^2
\end{equation}

Where $\mathcal{G}$ is the set of all the annotations for the image and $\hat{g}$ is the network prediction.

Before training, we perform data augmentation by translating, rotating and mirroring the images. For synthetic data, we also use the object's mask to replace the default background with different textures (cardboard, paper, wood, grass ...) to generate more variabilities.

\subsection{Cross-dataset evaluation}
\label{subsect:cross-dataset evaluation}

\begin{table}[]
\centering
\caption{Accuracy of the network trained on different datasets}
\label{table_2}
\setlength\tabcolsep{4pt}
\begin{tabular}{|c|c|c|c|}
\hline
\multicolumn{1}{|c|}{\multirow{2}{*}{\textbf{Training Dataset}}} & \multicolumn{2}{c|}{\textbf{Rectangle Metrics}} & 
\multicolumn{1}{c|}{\textbf{SGT}} \\ \cline{2-4} 
\multicolumn{1}{|c|}{} & \multicolumn{1}{c|}{Cornell} & \multicolumn{1}{c|}{Jacquard} & \multicolumn{1}{c|}{Jacquard} \\ \hline
Cornell & 86.88\% $\pm$ 2.57  & 54.28\% $\pm$ 1.22 & 42.76\% $\pm$ 0.91  \\
Jacquard (ours) & 81.92\% $\pm$ 1.95 & 74.21\% $\pm$ 0.71& 72.42\% $\pm$ 0.80 \\ \hline
\end{tabular}
\end{table}

This series of experiments aims to show that: 1) our Jacquard grasp dataset, despite being synthetic, can be used to train DNNs to predict grasp locations on real images; 2) The diversity of objects and grasp locations is important for a trained CNN to generalize on unseen objects. For this purpose, the Cornell dataset with its 885 RGB-D images on 240 objects and 8019 hand labeled grasp locations is used along with a portion of Jacquard which contains 15k RGB-D images on a selected 3k objects and 316k different grasp positions. To highlight 1), Alexnet is trained on Jacquard and tested on Cornell; for 2) it is trained on Cornell and tested on Jacquard. For comparison, we also display a baseline performance with Alexnet trained and tested on the same dataset, \textit{i.e.}, Cornell or Jacquard. For this purpose, we performed training and testing of the network with 5-fold cross validation for Cornell or Jacquard, leading to 5 variants of Alexnet with slightly different accuracies on each dataset. Each variant trained on Cornell (Jacquard, respectively) is then tested on the whole Jacquard dataset (Cornell, respectively) to evidence 1).

\hyperref[table_2]{Table II} summarizes the experimental results evaluated by both rectangle metrics and SGT criterion. As can be seen from \hyperref[table_2]{Table II}, when Alexnet is trained on our simulated Jacquard dataset and tested on Cornell, it achieves a grasp prediction accuracy of 81.92\% which is quite close to the baseline performance of 86.88\%. Furthermore, we also noticed that the networks trained on synthetic data tended to predict grasps which were visually correct despite being classified as wrong by the rectangle metrics. Typical examples are shown on the bottom line of \hyperref[misclassification]{Fig. 4}.

In contrast, when Alexnet is trained on Cornell and tested on Jacquard with a much wider diversity of objects and grasps, it depicts a grasp prediction accuracy of 54.28\% which records a performance decrease of 20 points in comparison with its baseline performance. As for the other training, part of this gap could be explained by the misclassifications of the rectangle metrics. However, this performance decrease is confirmed by our criterion based on simulated grasp trials (SGT): Alexnet trained on Cornell only displays a grasp prediction accuracy of 42.76\% which is 30 points behind the 72.42\% accuracy of the same CNN trained on Jacquard.


All these figures thus suggest that Jacquard can be used to train CNN, for an effective prediction of grasp locations. Furthermore, thanks to the diversity of objects and grasp locations, Jacquard enables a much better generalization skills of the trained CNN.



\subsection{Evaluation of grasp predictions using a real grasping robot}
\label{subsect:Evaluation of grasp predictions using a real grasping robot}

\begin{figure}
	\centering
	\def\svgwidth{0.45\textwidth}
	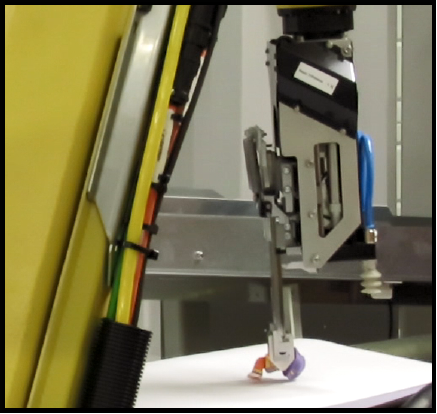
	\caption{Our physical setup to test grasp predictions. The camera is located above the grasping area.}
	\label{physical_setup}
\end{figure}

How good is a grasp predicted by a trained deep neural network, in real? To answer this question of possible reality gap, we used a parallel plate gripper mounted on a Fanuc's M-20iA robotic arm and a set of various objects. To ensure a wide variability in shapes, materials and scales, we used 15 everyday objects (toys and furnitures) and 13 industrial components. \hyperref[physical_setup]{Fig. 5} shows the robot performing a predicted grasp on one of the testing objects. Our criterion of a successful grasp was the same as in the simulator but this time using the aforementioned real grasping robot instead of the simulated one: the grasp of an object is considered successful only if the object is lifted, moved away and correctly dropped. For this test, we compared Alexnet trained on the Cornell dataset and the same network trained on a subset of 2k objects from the Jacquard dataset.

The experimental results show that the grasp predictor with Alexnet trained on the Jacquard dataset displays a grasp successful rate of 78.43\% which is even 6 points higher than the grasp accuracy displayed by Alexnet when it was trained and tested on the subset of 3k objects of Jacquard (see \hyperref[table_2]{Table II}) using the SGT criterion. This generalization skill of the trained grasp predictor can be explained by the large diversity of objects and grasp locations in the Jacquard dataset. For most of the failed cases, the grasp was not stable enough: the rectangle in the image was visually coherent and the object was successfully lifted but dropped during the movement of the robot.

\begin{figure}
	\centering
	\def\svgwidth{0.45\textwidth}
	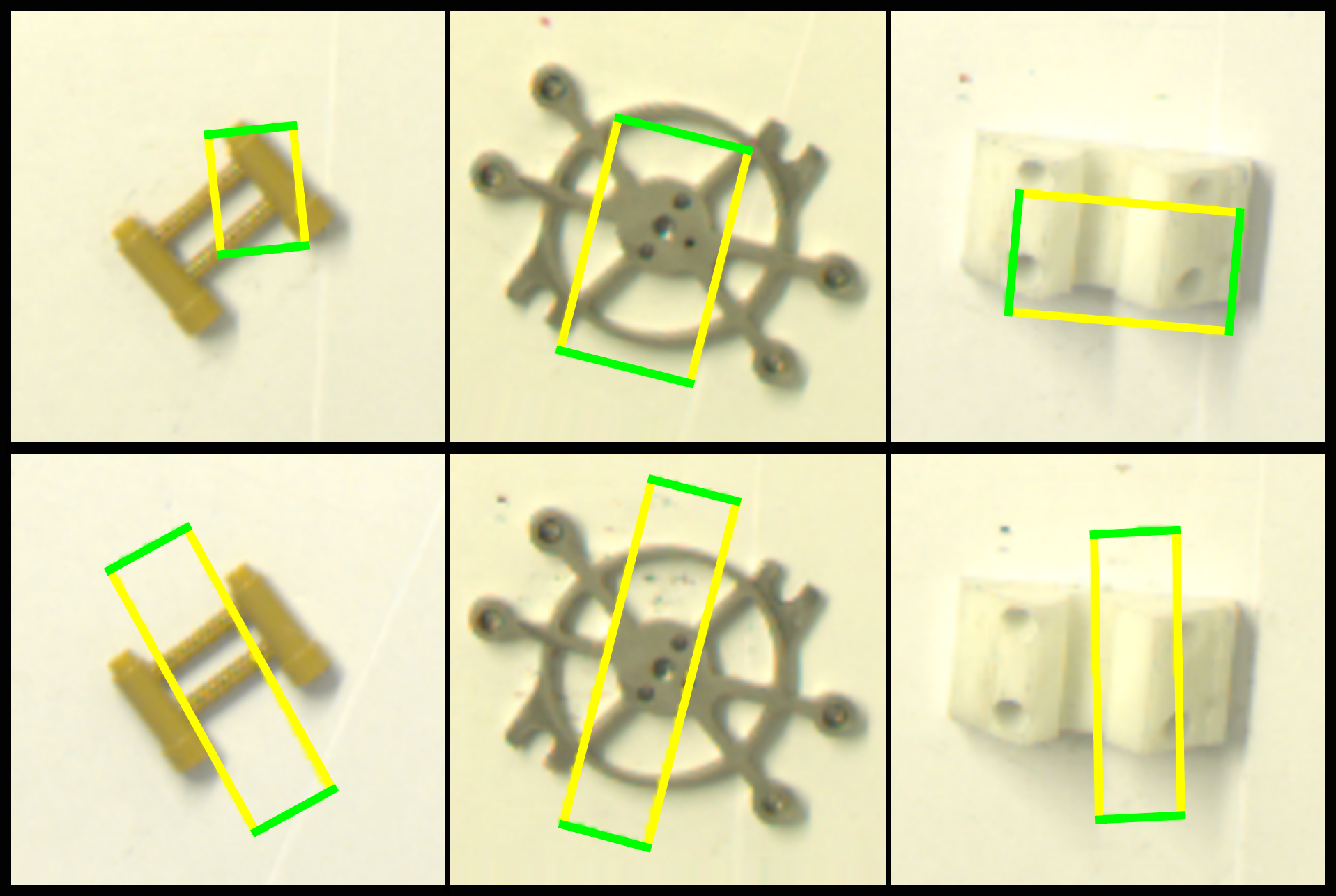
	\caption{Samples of grasp predictions on our real setup for the network trained on the Cornell dataset (top row) and the one trained on our synthetic Jacquard data (bottom row).}
	\label{real_grasps}
\end{figure}

Now with the the same network trained on Cornell, the robot succeeded only 60.46\% of the predicted grasps, mostly due to bad rectangle localization in the image. \hyperref[real_grasps]{Fig. 6} shows some examples of the objects for which the network trained on Cornell failed to predict a good grasp while the one trained on Jacquard succeeded.

\section{CONCLUSIONS}

In this work, we presented a method to generate realistic RGB-D data with localized grasp annotations from simulation. Using this method, we built a large scale grasp dataset with simulated data, namely Jacquard, and we successfully used it to train a deep neural network to predict grasp positions in images. The grasp predictor trained using Jacquard shows a much better generalization skill than the same network when trained with a small hand labeled grasp dataset. Our future work will focus on the quality assessment of grasp predictions and on extending this method to more complexe scenes, for example with multiple objects.





\section*{ACKNOWLEDGMENT}

This work was in part supported by the EU FEDER, Saint-Etienne M\'etropole and  R\'egion Auvergne-Rh\^{o}ne-Alpes fundings through the FUI PIKAFLEX project and in part by the French National Research Agency, l'Agence Nationale de Recherche (ANR), through the ARES labcom project under grant ANR 16-LCV2-0012-01.

\bibliographystyle{IEEEtran}
\bibliography{IEEEabrv,biblio}

\end{document}

%% file: Fig_Grasp_Notations.pdf_tex
\begingroup%
  \makeatletter%
  \providecommand\color[2][]{%
    \errmessage{(Inkscape) Color is used for the text in Inkscape, but the package 'color.sty' is not loaded}%
    \renewcommand\color[2][]{}%
  }%
  \providecommand\transparent[1]{%
    \errmessage{(Inkscape) Transparency is used (non-zero) for the text in Inkscape, but the package 'transparent.sty' is not loaded}%
    \renewcommand\transparent[1]{}%
  }%
  \providecommand\rotatebox[2]{#2}%
  \ifx\svgwidth\undefined%
    \setlength{\unitlength}{279.10158184bp}%
    \ifx\svgscale\undefined%
      \relax%
    \else%
      \setlength{\unitlength}{\unitlength * \real{\svgscale}}%
    \fi%
  \else%
    \setlength{\unitlength}{\svgwidth}%
  \fi%
  \global\let\svgwidth\undefined%
  \global\let\svgscale\undefined%
  \makeatother%
  \begin{picture}(1,0.89827041)%
    \put(0,0){\includegraphics[width=\unitlength,page=1]{Fig_Grasp_Notations.pdf}}%
    \put(0.68330499,0.48848336){\color[rgb]{0,0,0}\makebox(0,0)[b]{\smash{$\theta$}}}%
    \put(0,0){\includegraphics[width=\unitlength,page=2]{Fig_Grasp_Notations.pdf}}%
    \put(0.40838102,0.61718812){\color[rgb]{0,0,0}\makebox(0,0)[b]{\smash{$w$}}}%
    \put(0,0){\includegraphics[width=\unitlength,page=3]{Fig_Grasp_Notations.pdf}}%
    \put(0.18986573,0.19237269){\color[rgb]{0,0,0}\makebox(0,0)[b]{\smash{$h$}}}%
    \put(0.53002667,0.40363601){\color[rgb]{0,0,0}\makebox(0,0)[b]{\smash{$(x,y)$}}}%
    \put(0,0){\includegraphics[width=\unitlength,page=4]{Fig_Grasp_Notations.pdf}}%
  \end{picture}%
\endgroup%

%% file: Generation_Pipeline.pdf_tex
\begingroup%
  \makeatletter%
  \providecommand\color[2][]{%
    \errmessage{(Inkscape) Color is used for the text in Inkscape, but the package 'color.sty' is not loaded}%
    \renewcommand\color[2][]{}%
  }%
  \providecommand\transparent[1]{%
    \errmessage{(Inkscape) Transparency is used (non-zero) for the text in Inkscape, but the package 'transparent.sty' is not loaded}%
    \renewcommand\transparent[1]{}%
  }%
  \providecommand\rotatebox[2]{#2}%
  \ifx\svgwidth\undefined%
    \setlength{\unitlength}{537.19268181bp}%
    \ifx\svgscale\undefined%
      \relax%
    \else%
      \setlength{\unitlength}{\unitlength * \real{\svgscale}}%
    \fi%
  \else%
    \setlength{\unitlength}{\svgwidth}%
  \fi%
  \global\let\svgwidth\undefined%
  \global\let\svgscale\undefined%
  \makeatother%
  \begin{picture}(1,0.37364356)%
    \put(0,0){\includegraphics[width=\unitlength,page=1]{Generation_Pipeline.pdf}}%
    \put(0.07429393,0.01032211){\color[rgb]{0,0,0}\makebox(0,0)[b]{\smash{3D Models database}}}%
    \put(0,0){\includegraphics[width=\unitlength,page=2]{Generation_Pipeline.pdf}}%
    \put(0.23657514,0.07393869){\color[rgb]{0,0,0}\makebox(0,0)[b]{\smash{Scene creation}}}%
    \put(0,0){\includegraphics[width=\unitlength,page=3]{Generation_Pipeline.pdf}}%
    \put(0.59287313,0.00307589){\color[rgb]{0,0,0}\makebox(0,0)[b]{\smash{Rendering}}}%
    \put(0.60547323,0.19235107){\color[rgb]{0,0,0}\makebox(0,0)[b]{\smash{Grasps simulation}}}%
    \put(0,0){\includegraphics[width=\unitlength,page=4]{Generation_Pipeline.pdf}}%
    \put(0.92409624,0.18901206){\color[rgb]{0,0,0}\makebox(0,0)[b]{\smash{Successful grasps}}}%
    \put(0.23717996,0.22852788){\color[rgb]{0,0,0}\makebox(0,0)[b]{\smash{Random grasps}}}%
    \put(0,0){\includegraphics[width=\unitlength,page=5]{Generation_Pipeline.pdf}}%
    \put(0.42475611,0.03245948){\color[rgb]{0,0,0}\makebox(0,0)[b]{\smash{\shortstack{RGB\\image}}}}%
    \put(0.53493362,0.03237824){\color[rgb]{0,0,0}\makebox(0,0)[b]{\smash{\shortstack{True\\depth}}}}%
    \put(0.64498226,0.03237824){\color[rgb]{0,0,0}\makebox(0,0)[b]{\smash{\shortstack{Stereo\\depth}}}}%
    \put(0.75494028,0.02966593){\color[rgb]{0,0,0}\makebox(0,0)[b]{\smash{\shortstack{Object\\mask}}}}%
    \put(0,0){\includegraphics[width=\unitlength,page=6]{Generation_Pipeline.pdf}}%
  \end{picture}%
\endgroup%

%% file: missclassifications.pdf_tex
\begingroup%
  \makeatletter%
  \providecommand\color[2][]{%
    \errmessage{(Inkscape) Color is used for the text in Inkscape, but the package 'color.sty' is not loaded}%
    \renewcommand\color[2][]{}%
  }%
  \providecommand\transparent[1]{%
    \errmessage{(Inkscape) Transparency is used (non-zero) for the text in Inkscape, but the package 'transparent.sty' is not loaded}%
    \renewcommand\transparent[1]{}%
  }%
  \providecommand\rotatebox[2]{#2}%
  \ifx\svgwidth\undefined%
    \setlength{\unitlength}{664.53541872bp}%
    \ifx\svgscale\undefined%
      \relax%
    \else%
      \setlength{\unitlength}{\unitlength * \real{\svgscale}}%
    \fi%
  \else%
    \setlength{\unitlength}{\svgwidth}%
  \fi%
  \global\let\svgwidth\undefined%
  \global\let\svgscale\undefined%
  \makeatother%
  \begin{picture}(1,0.6689417)%
    \put(0,0){\includegraphics[width=\unitlength,page=1]{missclassifications.pdf}}%
  \end{picture}%
\endgroup%

%% file: Physical_setup.pdf_tex
\begingroup%
  \makeatletter%
  \providecommand\color[2][]{%
    \errmessage{(Inkscape) Color is used for the text in Inkscape, but the package 'color.sty' is not loaded}%
    \renewcommand\color[2][]{}%
  }%
  \providecommand\transparent[1]{%
    \errmessage{(Inkscape) Transparency is used (non-zero) for the text in Inkscape, but the package 'transparent.sty' is not loaded}%
    \renewcommand\transparent[1]{}%
  }%
  \providecommand\rotatebox[2]{#2}%
  \ifx\svgwidth\undefined%
    \setlength{\unitlength}{125.7030469bp}%
    \ifx\svgscale\undefined%
      \relax%
    \else%
      \setlength{\unitlength}{\unitlength * \real{\svgscale}}%
    \fi%
  \else%
    \setlength{\unitlength}{\svgwidth}%
  \fi%
  \global\let\svgwidth\undefined%
  \global\let\svgscale\undefined%
  \makeatother%
  \begin{picture}(1,0.94669483)%
    \put(0,0){\includegraphics[width=\unitlength,page=1]{Physical_setup.pdf}}%
  \end{picture}%
\endgroup%

%% file: real_robot.pdf_tex
\begingroup%
  \makeatletter%
  \providecommand\color[2][]{%
    \errmessage{(Inkscape) Color is used for the text in Inkscape, but the package 'color.sty' is not loaded}%
    \renewcommand\color[2][]{}%
  }%
  \providecommand\transparent[1]{%
    \errmessage{(Inkscape) Transparency is used (non-zero) for the text in Inkscape, but the package 'transparent.sty' is not loaded}%
    \renewcommand\transparent[1]{}%
  }%
  \providecommand\rotatebox[2]{#2}%
  \ifx\svgwidth\undefined%
    \setlength{\unitlength}{552.03537547bp}%
    \ifx\svgscale\undefined%
      \relax%
    \else%
      \setlength{\unitlength}{\unitlength * \real{\svgscale}}%
    \fi%
  \else%
    \setlength{\unitlength}{\svgwidth}%
  \fi%
  \global\let\svgwidth\undefined%
  \global\let\svgscale\undefined%
  \makeatother%
  \begin{picture}(1,0.67069924)%
    \put(0,0){\includegraphics[width=\unitlength,page=1]{real_robot.pdf}}%
  \end{picture}%
\endgroup%

%% file: root.bbl
\begin{thebibliography}{10}
\providecommand{\url}[1]{#1}
\csname url@rmstyle\endcsname
\providecommand{\newblock}{\relax}
\providecommand{\bibinfo}[2]{#2}
\providecommand\BIBentrySTDinterwordspacing{\spaceskip=0pt\relax}
\providecommand\BIBentryALTinterwordstretchfactor{4}
\providecommand\BIBentryALTinterwordspacing{\spaceskip=\fontdimen2\font plus
\BIBentryALTinterwordstretchfactor\fontdimen3\font minus
  \fontdimen4\font\relax}
\providecommand\BIBforeignlanguage[2]{{%
\expandafter\ifx\csname l@#1\endcsname\relax
\typeout{** WARNING: IEEEtran.bst: No hyphenation pattern has been}%
\typeout{** loaded for the language `#1'. Using the pattern for}%
\typeout{** the default language instead.}%
\else
\language=\csname l@#1\endcsname
\fi
#2}}

\bibitem{redmon2015real}
J.~Redmon and A.~Angelova, ``Real-time grasp detection using convolutional
  neural networks,'' in \emph{Robotics and Automation (ICRA), 2015 IEEE
  International Conference on}.\hskip 1em plus 0.5em minus 0.4em\relax IEEE,
  2015, pp. 1316--1322.

\bibitem{mahler2017dex}
J.~Mahler, J.~Liang, S.~Niyaz, M.~Laskey, R.~Doan, X.~Liu, J.~A. Ojea, and
  K.~Goldberg, ``Dex-net 2.0: Deep learning to plan robust grasps with
  synthetic point clouds and analytic grasp metrics,'' \emph{arXiv preprint
  arXiv:1703.09312}, 2017.

\bibitem{pinto2016supersizing}
L.~Pinto and A.~Gupta, ``Supersizing self-supervision: Learning to grasp from
  50k tries and 700 robot hours,'' in \emph{Robotics and Automation (ICRA),
  2016 IEEE International Conference on}.\hskip 1em plus 0.5em minus
  0.4em\relax IEEE, 2016, pp. 3406--3413.

\bibitem{levine2016learning}
S.~Levine, P.~Pastor, A.~Krizhevsky, J.~Ibarz, and D.~Quillen, ``Learning
  hand-eye coordination for robotic grasping with deep learning and large-scale
  data collection,'' \emph{The International Journal of Robotics Research},
  2016.

\bibitem{rodriguez2012caging}
A.~Rodriguez, M.~T. Mason, and S.~Ferry, ``From caging to grasping,'' \emph{The
  International Journal of Robotics Research}, vol.~31, no.~7, pp. 886--900,
  2012.

\bibitem{miller2004graspit}
A.~T. Miller and P.~K. Allen, ``Graspit! a versatile simulator for robotic
  grasping,'' \emph{IEEE Robotics \& Automation Magazine}, vol.~11, no.~4, pp.
  110--122, 2004.

\bibitem{miller2003automatic}
A.~T. Miller, S.~Knoop, H.~I. Christensen, and P.~K. Allen, ``Automatic grasp
  planning using shape primitives,'' in \emph{Robotics and Automation, 2003.
  Proceedings. ICRA'03. IEEE International Conference on}, vol.~2.\hskip 1em
  plus 0.5em minus 0.4em\relax IEEE, 2003, pp. 1824--1829.

\bibitem{bohg2010learning}
J.~Bohg and D.~Kragic, ``Learning grasping points with shape context,''
  \emph{Robotics and Autonomous Systems}, vol.~58, no.~4, pp. 362--377, 2010.

\bibitem{goldfeder2009columbia}
C.~Goldfeder, M.~Ciocarlie, H.~Dang, and P.~K. Allen, ``The columbia grasp
  database,'' in \emph{Robotics and Automation, 2009. ICRA'09. IEEE
  International Conference on}.\hskip 1em plus 0.5em minus 0.4em\relax IEEE,
  2009, pp. 1710--1716.

\bibitem{lecun2015deep}
Y.~LeCun, Y.~Bengio, and G.~Hinton, ``Deep learning,'' \emph{Nature}, vol. 521,
  no. 7553, pp. 436--444, 2015.

\bibitem{oberlin15scale}
J.~Oberlin, M.~Meier, T.~Kraska, and S.~Tellex, ``{Acquiring Object Experiences
  at Scale},'' in \emph{{AAAI-RSS Special Workshop on the 50th Anniversary of
  Shakey: The Role of AI to Harmonize Robots and Humans}}, 2015, blue Sky
  Award.

\bibitem{johns2016deep}
E.~Johns, S.~Leutenegger, and A.~J. Davison, ``Deep learning a grasp function
  for grasping under gripper pose uncertainty,'' in \emph{Intelligent Robots
  and Systems (IROS), 2016 IEEE/RSJ International Conference on}.\hskip 1em
  plus 0.5em minus 0.4em\relax IEEE, 2016, pp. 4461--4468.

\bibitem{jiang2011efficient}
Y.~Jiang, S.~Moseson, and A.~Saxena, ``Efficient grasping from rgbd images:
  Learning using a new rectangle representation,'' in \emph{Robotics and
  Automation (ICRA), 2011 IEEE International Conference on}.\hskip 1em plus
  0.5em minus 0.4em\relax IEEE, 2011, pp. 3304--3311.

\bibitem{lenz2015deep}
I.~Lenz, H.~Lee, and A.~Saxena, ``Deep learning for detecting robotic grasps,''
  \emph{The International Journal of Robotics Research}, vol.~34, no. 4-5, pp.
  705--724, 2015.

\bibitem{shapenet2015}
A.~X. Chang, T.~Funkhouser, L.~Guibas, P.~Hanrahan, Q.~Huang, Z.~Li,
  S.~Savarese, M.~Savva, S.~Song, H.~Su, J.~Xiao, L.~Yi, and F.~Yu,
  ``{ShapeNet: An Information-Rich 3D Model Repository},'' Stanford University
  --- Princeton University --- Toyota Technological Institute at Chicago, Tech.
  Rep. arXiv:1512.03012 [cs.GR], 2015.

\bibitem{savva2015semgeo}
M.~Savva, A.~X. Chang, and P.~Hanrahan, ``{Semantically-Enriched 3D Models for
  Common-sense Knowledge},'' \emph{CVPR 2015 Workshop on Functionality,
  Physics, Intentionality and Causality}, 2015.

\bibitem{coumans2017}
E.~Coumans and Y.~Bai, ``pybullet, a python module for physics simulation for
  games, robotics and machine learning,'' \url{http://pybullet.org/},
  2016--2017.

\bibitem{blender2016}
\BIBentryALTinterwordspacing
{Blender Online Community}, \emph{Blender - a 3D modelling and rendering
  package}, Blender Foundation, Blender Institute, Amsterdam, 2016. [Online].
  Available: \url{http://www.blender.org}
\BIBentrySTDinterwordspacing

\bibitem{hirschmuller2008stereo}
H.~Hirschmuller, ``Stereo processing by semiglobal matching and mutual
  information,'' \emph{IEEE Transactions on pattern analysis and machine
  intelligence}, vol.~30, no.~2, pp. 328--341, 2008.

\bibitem{bregier2017symmetry}
R.~Br{\'e}gier, F.~Devernay, L.~Leyrit, and J.~L. Crowley, ``Symmetry aware
  evaluation of 3d object detection and pose estimation in scenes of many parts
  in bulk,'' in \emph{Proceedings of the IEEE International Conference on
  Computer Vision}, 2017, pp. 2209--2218.

\bibitem{vhacd16}
K.~Mamou, ``Volumetric hierarchical approximate convex decomposition,'' in
  \emph{Game Engine Gems 3}, E.~Lengyel, Ed.\hskip 1em plus 0.5em minus
  0.4em\relax A K Peters, 2016, pp. 141--158.

\bibitem{krizhevsky2012imagenet}
A.~Krizhevsky, I.~Sutskever, and G.~E. Hinton, ``Imagenet classification with
  deep convolutional neural networks,'' in \emph{Advances in neural information
  processing systems}, 2012, pp. 1097--1105.

\bibitem{imagenet_cvpr09}
J.~Deng, W.~Dong, R.~Socher, L.-J. Li, K.~Li, and L.~Fei-Fei, ``{ImageNet: A
  Large-Scale Hierarchical Image Database},'' in \emph{CVPR09}, 2009.

\end{thebibliography}
